# Learned and Controlled Autonomous Robotic Exploration in an Extreme, Unknown Environment

Frances Zhu
Cornell University
452 Upson Hall
Ithaca, NY 14853
fz55@cornell.edu

D. Sawyer Elliott
Cornell University
452 Upson Hall
Ithaca, NY 14853
dse44@cornell.edu

ZhiDi Yang
Cornell University
Ithaca, NY 14853
zy337@cornell.edu

Haoyuan Zheng
Cornell University
Ithaca, NY 14853
hz463@cornell.edu

*Abstract*— Exploring and traversing extreme terrain with surface robots is difficult, but highly desirable for many applications, including exploration of planetary surfaces, search and rescue, among others. For these applications, to ensure the robot can predictably locomote, the interaction between the terrain and vehicle, terramechanics, must be incorporated into the model of the robot's locomotion. Modeling terramechanic effects is difficult and may be impossible in situations where the terrain is not known *a priori*. For these reasons, learning a terramechanics model online is desirable to increase the predictability of the robot's motion. A problem with previous implementations of learning algorithms is that the terramechanics model and corresponding generated control policies are not easily interpretable or extensible. If the models were of interpretable form, designers could use the learned models to inform vehicle and/or control design changes to refine the robot architecture for future applications. This paper explores a new method for learning a terramechanics model and a control policy using a model-based genetic algorithm. The proposed method yields an interpretable model, which can be analyzed using preexisting analysis methods. The paper provides simulation results that show for a practical application, the genetic algorithm performance is approximately equal to the performance of a state-of-the-art neural network approach, which does not provide an easily interpretable model.

## I. Introduction

Exploring environments with extreme terrain is difficult for robotic systems. However, for many applications, such as search and rescue missions and planetary exploration, effectively exploring extreme terrain is crucial. One aspect that complicates exploration of extreme terrain is that the dynamics interaction between the robot and terrain, the terramechanics, are often not known accurately *a priori*. When traversing extreme terrain, not accurately modeling the terramechanics inhibits the capability of robotic system to predictably locomote. As the terrain becomes more extreme or as the robot's motion becomes more agile, the adverse effect of poorly modeled terramechanics is even more significant because difficult-to-model nonlinearities and discontinuities from the environment's forces affect the body more significantly. Thus, to traverse extreme terrain effectively and predictability, the terramechanic effects must be incorporated into the control system of the robot. To incorporate terramechanic effects into the control system when the terramechanics are not known *a priori,* the terramechanics must be learned online.

Many methods have been used in past research to develop terramechanics models online. A promising method is to use reinforcement learning to efficiently learn a general relationship between the dynamics of the body and the environment, which includes the terramechanic effects. Learning a black box model using neural networks (NNs) has been shown to increase the ability of the robot to traverse an unknown environment effectively, but is difficult to interpret by designers [1], [2], [3], [4], [5]. For many practical applications, users need transparency to better understand and refine robotic systems. For example, if the learned dynamic interaction between the robot and environment yields a reduced proportional relationship between expected acceleration and wheel speed, viscous drag can be identified. With a learned dynamic relationship, scientists may induce physical traits of the environment, which contributes two-fold to the planetary science field. If the learned controller is in standard nonlinear control form, the model can be analyzed during or after operation using standard analysis techniques, informing control design and/or vehicle design improvements for future missions.

In this paper, we propose a control system that learns an interpretable model of the terramechanics online and computes an optimal controller that enables accurate trajectory tracking. Specifically, this paper proposes a genetic learning algorithm (GA), which learns a model of the terramechanics along with an optimal controller in standard nonlinear control form.

In simulation, the controller is implemented on a vehicle with Ackerman steering, enabling the vehicle to accurately track a trajectory through an environment with unknown terramechanics. The simulation results show that the proposed control, using a genetic algorithm, enables the vehicle to track a trajectory with approximately the same performance as a state-of-the-art NN, while also offering an interpretable terramechanics model stemming from fundamental physics.

## II. Background

Modeling terramechanics is a large and active field of study. Past research has explored how to model motion through extreme terrains like sand, mud, ice, and how to incorporate terramechanics to create controllers that track trajectories effectively [6], [7], [8]. However, these methods assume that the terramechanics are known *a priori,* which as discussed before, is not always the case, especially for applications of space exploration. Due to the inaccessibility of space and technological immaturity of reinforcement learning algorithms, previous work draws from the robotics community that focus on predominantly ground experiments.

For cases where the terramechanics are not known accurately *a priori,* methods for learning the terramechanics and controls have been proposed previously. For example, unsupervised learning has been used to classify the sliding events of discrete rovers, which enables more accurate tracking of trajectories [9]. Reinforcement learning with neural networks has also been applied to learning terramechanics models and controllers for autonomous robots, including drifting, walking on extreme terrain, traversing over obstacles, among others [1], [2], [3], [4], [5]. However, none of these methods output an easily interpretable model of the terramechanics or controller, which makes analysis of the resulting model difficult.

Genetic algorithms have been used to effectively locomote robots in a variety of environments both aquatic and terrestrial [10], [11]. This previous work focuses on the morphology of the robot that best achieves locomotion, not learning a dynamics model. To address system identification, another divergent line of research learns a symbolic expression of a dynamics model with control input that most disagrees with candidate physical models in a controlled environment [12], [13]. Implementing control input with the most disagreement risks of immobilization for an extraplanetary rover operating outside the confines of a controlled laboratory environment. Aerospace applications desire robustness in autonomous operations, which involve guarantees and predictive confidence.

### III. CONTROL DESIGN

To enable exploration of extreme terrain and learn an interpretable model, a two-part controller is proposed. The first part of the controller is a genetic algorithm, which learns a dynamic model, including terramechanic effects, and a controller in the form of an adaptive, linear gain matrix. The second part of the controller is a baseline controller, which uses the commonly employed pure pursuit method, to roughly track the trajectory such that the learning controller can gather enough data to learn a dynamics model and a control policy. The baseline controller could be removed if the learning algorithm was trained using a dynamic simulation. However, the training is only as accurate as the modeled system structure. Thus, to accurately train the learning algorithm, an accurate model of the system would need to be developed, which is difficult and potentially impossible for unknown terrains. The dynamic model's static structure offers limitations in behavior that could offer safety in the form of guarantees.

#### A. Reinforcement Learner Specification & Design

The proposed learning method uses a genetic algorithm evolving a multivariable, nonlinear model approximation. To achieve efficient computation and to ensure the structure is interpretable, the genetic algorithm assumes a static model structure. The parameters in the structure are evolved, or estimated, to provide a candidate physics model and create a control policy.

#### B. Implementation

Many methods exist for implementing genetic algorithms. A basic genetic algorithm includes population initialization, fitness evaluation, reproduction, crossover, and mutation [14]. The important characteristics of a genetic algorithm are chromosome specification, evolution parameters, and fitness functions. The chosen method for the proposed controller is discussed below.

Two populations describe candidates for the dynamic model parameters and optimal control policy parameters. Parameters are analogously called chromosomes in the context of genetic algorithms. For unknown terramechanics, length of the chromosome for the dynamic parameters is determined by the complexity of the chosen terramechanics model, given in Eq. (1). The dynamics model needs a dynamic parameter chromosome string, $\theta_d$, defined by a number $m_d$ parameters in which each component is $p_1$ to $p_{m_d}$. These parameters represent necessary coefficients in the dynamics model expression, like scalars and biases, capturing a number of physical effects, like static or sliding friction. For the optimal control population, the chromosome length is dependent on the complexity of the chosen control policy. The control policy needs a control parameter chromosome string, $\theta_K$, defined by a number $n_d$ parameters in which each component is $k_1$ to $k_{n_d}$, given in Eq. (2). The components resemble gain matrix values but reshaped into vector form, instead of the original matrix dimension. Each population evolves, guided by user-defined fitness metrics.

$$\theta_d = [p_1 \dots p_{m_d}] \quad (1)$$

$$\theta_K = [k_1 \dots k_{n_d}] \quad (2)$$

The two metrics for fitness evaluation of each population are prediction error $Q$ and tracking error $C$. The prediction error is the squared difference between measured and predicted states. The predication error is dotted with a weight vector $w_s$, shown in Eq. (4). Given the measured state of the $m^{th}$ previous timestep $s_{k-m}$ and the current timestep $s_k$, the prediction error is the error between the current measurement $s_k$ and the propagated state from the previous measurement $\hat{s}_k$. To propagate the vehicle's next dynamic states, each population member's dynamic parameters $\theta_{d,i}$ and the optimal actions taken since $k-m$ timesteps ago $a^*_{k-m}$, are injected into the prescribed dynamics model including terramechanics $f(\cdot)$, given in Eq. (3) and visually depicted in Figure 1. The fittest dynamic parameters are those that accurately reflect the physical system. $w_s$ can be modified depending on what portions of the predicated state the designer is more concerned about predicting accurately.

$$\hat{s}_{k,i} = f(\theta_{d,i}, s_{k-m}, a^*_{k-m}) \quad (3)$$

$$Q_i = w_s \cdot |(s_k - \hat{s}_{k,i})|^2 \quad (4)$$

$$C_i = w_r \cdot |(r_{k+n} - s'_{k+n,i})|^2 + w_k \cdot |\theta_{K,i}|^2 \quad (7)$$

The next generation is produced from the ranked population with probabilistic sampling. Parents are sampled from the ranked population with a standard Gaussian distribution, of which the fittest individuals are selected most often. With a crossover rate $C_r$, children are reproduced from the parent population by crossover from the two parental chromosomes. Every child's resultant chromosome is additionally mutated. Only the top $C_f$ members of the previous generation survive into the next generation. Finally, $C_n$ members enter the new generation to ensure the optimization process is adapting with the system, described in the next subsection.

The genetic algorithm progresses by reentering a loop to evaluate this new generation, which continues the evolution process. Allowing the system to implement the learned control input from the very start of the learning process could potentially be dangerous as the dynamics model has not been validated rigorously with enough measurements. The learner accumulates measurements and refines both the dynamics model parameters and control policy parameters until reaching a certain prediction and tracking error threshold, ensuring that the next state does not stray far from the reference state. Once that threshold is met, the learned control policy is run in the system's forefront.

The learned dynamics model and control policy adapt as information is gathered, differing from a system in which a dynamics model and control policy are specified *a priori*. The latter system does not have the opportunity to update, likely resulting in suboptimal trajectory tracking. The learned system offers two main advantages: accuracy and adaptability. The dynamics model incorporates the approximate terramechanics of the current terrain, likely offering more accurate trajectory tracking compared to a terramechanics model specified *a priori*. Additionally, the adaptive nature of the dynamics model extends to terrains with different properties, such as ice, steep slopes, and mud, thus unexpected terrain can be effectively traversed.

*1) Underdetermined System Identification*

Domain knowledge is critical to form a minimal formalization both in structure and parameters. A comprehensive model precisely characterizes a system but requires more system parameters, which increases evaluation computation and convergence time. Machine learning techniques leverage quick iteration and immense computation power by implementing a minimum description of the system [15].

The implemented dynamics model and control policy as proposed both suffer from being underdetermined. Consequently, the dynamic parameter and control parameter populations are at risk of prematurely converging to a local well. The dynamics model intentionally does not fully capture the system's terramechanics behavior but instead simplifies the model to reduce computation time in the algorithm. The control policy has a consistent, nonlinear mapping from state

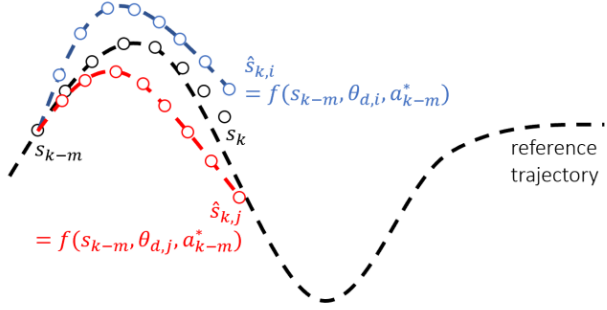

**Figure 1: Propagation of vehicle state using optimal actions $a^*_{k-m}$ and two population member's dynamic parameters $\theta_{d,i}$ and $\theta_{d,j}$**

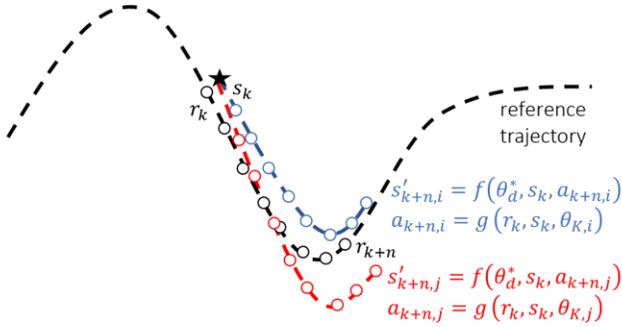

**Figure 2: Propagation of vehicle state using optimal dynamic parameters $\theta^*_d$ and two population member's control parameters $a_{k+n,i}$ and $a_{k+n,j}$**

Tracking error is the sum of error between the next to $n$ projected trajectory waypoints and velocities to the projected state $s'_{k+n}$, propagating a simulation with the optimal dynamic parameters $\theta^*_d$, the prescribed dynamics model including terramechanics $f(\cdot)$, and candidate control actions $a_{k+n,i}$, given in Eq. (5) and visually depicted in Figure 2. The selected control policy $g(\cdot)$ generates candidate control actions from the state $s_k$, a reference state $r_k$, and the candidate control parameters $\theta_{K,i}$ evolving in the genetic algorithm, given in Eq. (6). The final control fitness function is a weighted sum of squared error between the reference and projected states and squared weighted penalty of gain values, shown in Eq. (7). The squared weighted penalty of gain values is included to ensure the gains do not become so large that the system will become unstable. The fittest control parameters drive the system to the desired trajectory and velocity. The two populations are sorted from the most fitt members to the least fit members. From these two ranked populations, the next generation of each population is created using reproduction, crossover, and mutation.

$$s'_{k+n,i} = f(\theta^*_d, s_k, a_{k+n,i}) \quad (5)$$

$$a_{k+n,i} = g(r_k, s_k, \theta_{K,i}) \quad (6)$$

to input due to complex, unmodeled hardware effects but may be characterized locally in a linear mapping.

A new member is injected into the population at every generation to ensure that the genetic algorithm populations adapt locally with the system state. The dynamic parameter population receives a randomly generated member from the entire parameter space to guarantee a globally-scoped search. The control parameter population receives an inverse model mapping representative of the system within the immediate past timestep horizon of $h$, a local approximation. The inverse model is generated from vectors of previous control inputs $A_k$, corresponding state error $E_k$, and relevant system parameters $P_k$, given in Eq. (8) and Eq. (9). The newest control parameter member $\theta_{K,l}$ is the linear least squared error local approximation of the nonlinear control model, given in Eq. (10). The constant presence of this local approximation offers the genetic algorithm to adapt with the time-varying system if the evolved solutions do not track as well.

$$A_k = \begin{bmatrix} a_{k-h-1} \\ \vdots \\ a_{k-1} \end{bmatrix} \quad (8)$$

$$E_k = \begin{bmatrix} r_{k-h} - s_{k-h} \\ \vdots \\ r_k - s_k \end{bmatrix} \quad (9)$$

$$\theta_{K,l} = [E_k \quad P_k]^\dagger A_k \quad (10)$$

*C. Baseline Controller*

To enable the GA to learn effectively, the baseline controller is used to coarsely to track the trajectory. The baseline controller we propose is broken up into two sections: the velocity-tracking controller and the path-tracking controller. The velocity-tracking controller attempts to track the desired velocity profile of the trajectory. The velocity-tracking base controller is a proportional controller, as shown in Eq. (11), where $V_d$ is the desired velocity in the $\hat{b}_1$ direction of the car as described in the appendix, $V$ is the current velocity of the car, $K_p$ is a user-defined gain that is tuned on the physical system, and $C_V$ is the commanded wheel speed. The value of $K_p$ does not need to be fine-tuned, because after the learner gathers an appropriate amount of data, poor tuning will no longer affect the performance of the vehicle.

$$C_v = K_p(V_d - V \cdot \hat{b}_1) \quad (11)$$

The path-tracking controller attempts to track the path of the trajectory. The path-tracking base controller is a pure-pursuit controller. Pure-pursuit controllers are a common control strategy to enable a robot with Ackerman steering to track paths. The equation describing the pure-pursuit controller is shown in Eq. (12), where $L$ is the length between the front wheels and the real wheels, $L_d$ is a look-ahead gain, $\alpha$ is the path intersection angle, and $\phi$ is the computed steering angle, as discussed in [16].

$$\phi = \tan^{-1}\left(\frac{2L\sin(\alpha)}{L_d}\right) \quad (12)$$

## IV. EXPERIMENT

An experiment is run to determine if the proposed genetic learning algorithm control meets two main goals. The first goal is to verify if the GA can learn a dynamics model and controller in standard form when applied to a practical application. The second goal is to determine if the performance of the GA is approximately the same as a state-of-the-art NN approach. We hypothesize, that the two should produce approximately equal tracking performance. The performance metrics are error in trajectory tracking, convergence time, and algorithm computation time at every timestep. The metric used to determine how accurately a trajectory is tracked is shown in Eq. (14), where $r^{T/b}$ is the distance to the nearest portion of the trajectory from the car, $T_c$ is the time of convergence, and $T_f$ is the time to complete ten laps. The computation time at each timestep is measured using the algorithm environment's stopwatch timer.

$$J = \int_{T_c}^{T_f} |r^{T/b}| \, dt + \int_{T_c}^{T_f} (V_d - V \cdot \hat{b}_1) \, dt \quad (13)$$

The convergence time is how long it takes for the controller to converge to the trajectory and continue to track the trajectory accurately and repeatedly. Convergence time is determined by user intuition.

To test the genetic algorithm, a simulation of an Ackerman steering vehicle on a low friction inclined surface is used. The simulation is designed to mimic extreme terrains comprised of surfaces that are inclined and/or do not perfectly constrain the wheel's motion, such as sandy inclines or fine loose rubble, among others. The terramechanics of sand or fine loose rubble are different from a slippery incline and are more complicated. However, a slippery slope acts as a simplified test to understand the potential performance of the controller. The dynamics that are implemented into the simulation are derived below in the Appendix. To increase the validity of the simulation, noise is added to all states and an estimator is used to determine the state of the vehicle from only position orientation and time measurements. The noise is specified in Table 1, along with the parameters for the simulation environment and vehicle.

**Table 1: Parameters for the environment and vehicle**

| Parameter | Value |
| --- | --- |
| Dynamic parameter $\mu_s$ lateral wheel slip friction | 5 |
| Dynamic parameter $\mu_w$ forward wheel slip friction | 1 |
| Trajectory sloped surface angle $\delta$ | 30° |
| Vehicle mass | 1 kg |
| Vehicle wheel radius | 0.10 m |
| Distance from vehicle center of mass to | 0.16 m |
| center of rear wheel axle | |
| Measurement sampling time and control update rate | 0.2 sec |
| Measurement Gaussian position noise ($1\sigma$) | 1.3e-4 m |
| Measurement Gaussian rotation noise ($1\sigma$) | 0.83e-4 rad |

### A. Test Procedure

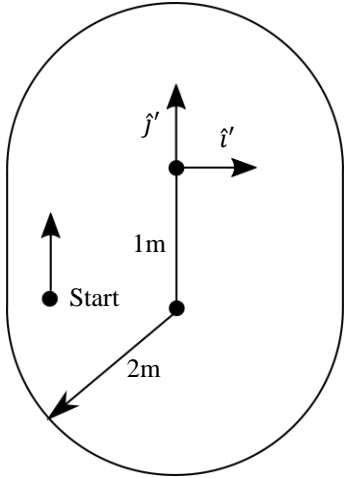

**Figure 3: Trajectory for the car to track**

To test the car, a predefined closed trajectory is specified for the RC car to track on the slippery slope. The trajectory remains constant throughout the test. Each controller is tested with the same starting configuration, which is consistently displaced from the reference trajectory. The car is commanded to track the trajectory, which is done at first using the baseline controller. Once the learned controller converges to a solution that meets a user-defined performance criterion, the trajectory is tracked using the learned controller. The desired tracking velocity is 0.2 m/s. The trajectory is shown in Figure 1.

### B. Genetic Algorithm Implementation

The dynamic model used for the GA is described below in the Appendix. From that derivation, the dynamic model parameter population members have chromosome of length two: friction coefficients, $\mu_s$ and $\mu_w$, describing lateral wheel slip and wheel slip in the direction of wheel velocity.

The form of the controller is a time-varying gain matrix. The action $a$ at every timestep is given by Eq. (14), where $\phi$ is the steering angle and $\omega_w$ is the rotation rate of the wheels. The control gain matrix is populated with the parameters in the control parameter chromosome, given by Eq. (15). The gain matrix maps the path intersection angle, error in velocity, and the estimated slope of the surface $\delta$ to the control actions $a$, given in Eq. (16).

$$a = [\phi \; \omega_w]' \quad (14)$$

$$K = \begin{bmatrix} \theta_{K,1} & \theta_{K,2} & \theta_{K,3} \\ \theta_{K,4} & \theta_{K,5} & \theta_{K,6} \end{bmatrix} \quad (15)$$

$$a = K[\alpha \; \Delta V \; \delta] \quad (16)$$

The fitness functions for each population are shown in Eq. (4) and (7), of which the specific weight matrices the remaining parameters for the GA are shown in Table 2.

**Table 2: Hyperparameters for evolution process**

| Parameter | Value |
|---|---|
| Steps compared for dynamics evaluation ($n$) | 1 |
| Steps compared for control evaluation ($m$) | 2 |
| Crossover Rate ($C_r$) | .67 |
| Size of dynamic population | 8 |
| Size of control population | 8 |
| Number of breeders ($C_f$) | 3 |
| Number of new members ($C_n$) | 1 |
| Weight of prediction vector $w_s$ | [$10^3$ $10^3$ 0 0 180/pi 0] |
| Weight of tracking vector $w_r$ | [1 1 0.01 0 0 0] |
| Weight control gain vector $w_K$ | [0 $10^{-7}$ 0 0 0 $10^{-7}$ 0] |

### C. Implementation of Neural Network for Comparison

The genetic algorithm is compared to a supervised neural network controller. The idea of neural network controller was first introduced by Demetri Psaltis et al. [17], in which an architecture is proposed for a general learning process. The idea was further developed by Tomochika el al. [1], in which a neural network tracks a trajectory with unstructured uncertainty. This supervised neural network approach builds on the referenced work. In this approach, the neural network is used as a function approximator to the cost function, more formally described in Eq. (17),

$$\hat{f}(s,a) \approx C(s,a) \quad (17)$$

where $s$ stands for the state and $a$ stands for the action. In this problem, to simplify the learning process, the state of vehicle is chosen as follows in Eq. (18):

$$s = [\varphi \; d \; \rho] \quad (18)$$

$\varphi$ : the angle between the body of the vehicle and the tangent of the target trajectory.
$d$ : the shortest distance from the vehicle to the trajectory.
$\rho$ : the angle between the body of the vehicle and the original point of axes.

The cost value describes how well the vehicle near the desired trajectory. Given the state value, the cost value of this state is explicitly defined in Eq. (19).

$$C = d^2 + k \cdot \rho^2 \quad (19)$$

The learning process consists of two stages. On the first stage, vehicle is controlled to do a random walk strategy to fully explore the target environment. In this process, all states, actions, and cost values are collected as training data. Using these data as training data, we train a neural network with three layers as the function approximator. After the first stage in the learning process, a cost function is learned, which could be used to develop a control strategy. At each state, a unique action could be selected to minimize the cost value based on the neural network. However, this process is time-consuming. Then, during the second learning stage, another neural network is used to directly describe the control strategy. The input of this neural network is the state and the output of the neural network is the action, given in Eq. (20).

$$a = \hat{g}(s) \tag{20}$$

Training data in this learning process is generated by running the first neural network. After the second learning process, the neural network is ready to be used as a controller.

This neural network controller doesn't use any dynamic model information and therefore is a model-free method, compared with the genetic algorithm. Due to limited space and time of training, the vehicle is subject to easily crash into the wall in the first learning stage. To address this problem, the first learning stage is implemented on the simulator.

## V. RESULTS

Both the learned controller and trained neural network tracked the desired trajectory better than the baseline controller, as shown in Figure 4. The tracking error integrated across ten laps after convergence is shown in Table 3 and the component error time history is depicted in Figure 5 and Figure 6. The total computation time to finish ten laps including the time to converge is shown in Table 4. The average computation time, along with a standard deviation, is reported in Table 5 with the computation at every timestep depicted in Figure 7. The convergence time for both methods are reported in Table 6 and depicted in Figure 8.

**Table 3: Total tracking error comparison across ten laps**

|  | Baseline | Learner | Neural Network |
|---|---|---|---|
| $J(r)$ [m] | 648 | 34 | 78 |
| $J(V)$ [m/s] | 1393 | 44 | 23 |
| $J_{tot}$ | 2041 | 78 | 101 |

**Table 4: Total computation time comparison over ten laps**

|  | Baseline | Learner | Neural Network |
|---|---|---|---|
| time | 348.7 s | 1244.1 s | 869 s |

**Table 5: Average computation time comparison on a dell Xenon desktop in MatLab**

|  | Baseline | Learner | Neural Network |
|---|---|---|---|
| mean | 35 ms | 330 ms | 341 ms |
| std | 15.4 ms | 123 ms | 26 ms |

**Table 6: Convergence time comparison**

|  | Baseline | Learner | Neural Network |
|---|---|---|---|
| control | NA | 217 s | 200 s |
| dynamics | NA | 40 – 130 s | NA |

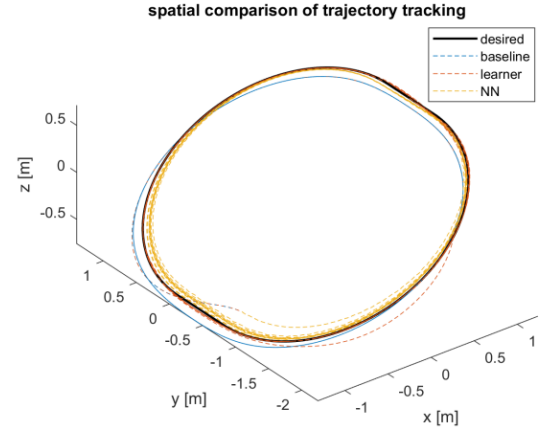

Figure 4: Trajectory comparison of baseline, learner, and neural network overlaid on desired trajectory

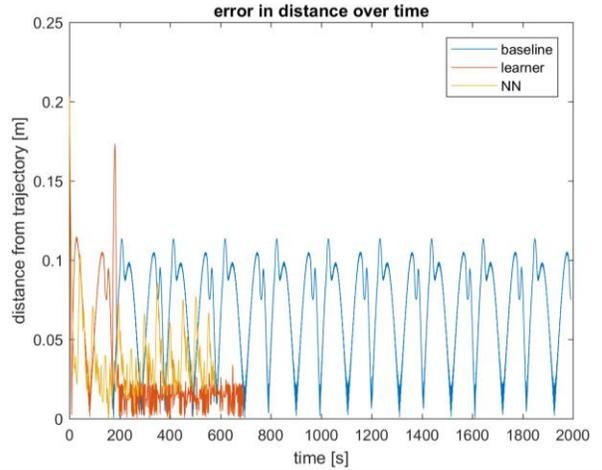

Figure 5: Time history of distance error comparison between baseline, real-time learner, and supervised neural network

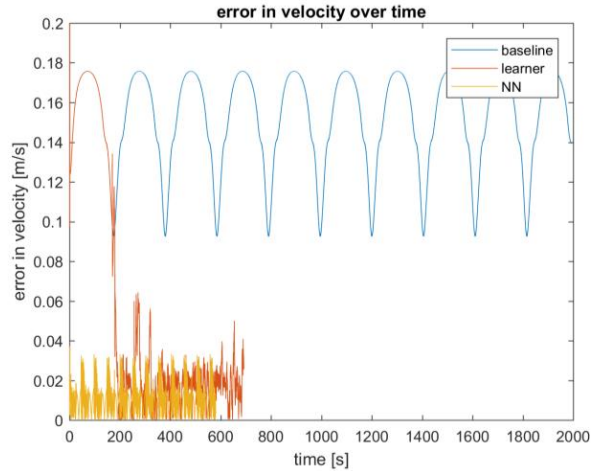

Figure 6: Time history of velocity error comparison between baseline, real-time learner, and supervised neural network

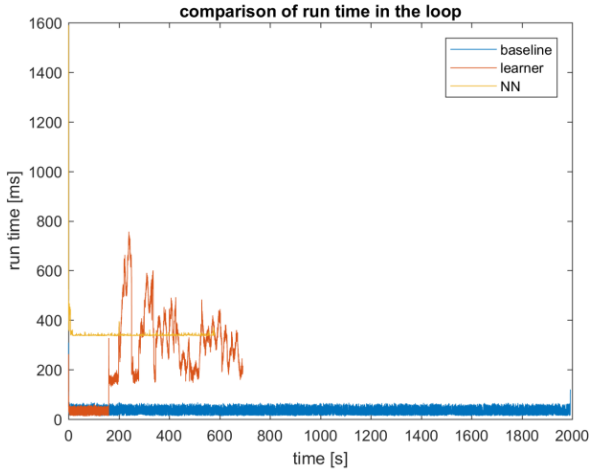

**Figure 7: Comparison of computation per evaluation loop over time**

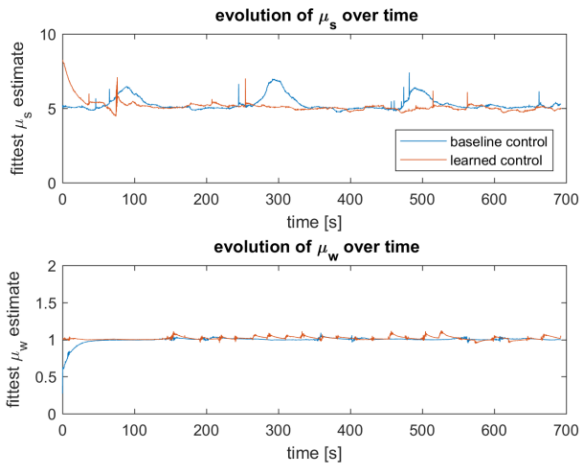

**Figure 8: Evolution of dynamic parameters over time**

From Figure 6, the GA converges to the real $\mu$ values with very little error. The noise that is seen in the estimate is due to the added noise in the measurements. The error for the controller also converges to an approximately steady value. Again, the variations are due to the injected noise. From Table 5, the average computation time is relatively small, making it possible to implement this method onto physical systems. From these results, the GA is capable of learning dynamics parameters for the simulated RC car, which supports the hypothesis that the method can be implemented on a practical system.

Also, from the results above the GA performs similarly to the NN in tracking error, average computation time, and convergence time. There are no appreciable differences in the reported performance metrics between the two methods, which supports our hypothesis, that the GA has similar performance to the state-of-the-art NN approaches, while providing an interpretable model.

## VI. LIMITATIONS

One major limitation of the proposed learned method is model bias due to underfitting, which results from some dynamic terms being excluded in the model structure. Model bias is caused by assuming a model structure that does not sufficiently describe the system. The model bias can be reduced by enabling the model structure to change. For example, the friction can be represented as a common friction model summed with a polynomial with varying order and coefficients. Enabling the GA to vary the order and coefficients enables more complicated friction models to be approximated. Similarly, polynomials can also be used to incorporate more complex dynamic effects or control methods.

Another limitation of the work is that the method proposed likely will not perform well when the dynamic parameters or optimal control parameters vary rapidly with time because the GA will likely not converge on rapidly changing parameters. For the simulation, the time varying effects of the control parameters were slow, on the order of a lap, and the true dynamic parameters were constant, so the GA could continuously update the parameters. However, for other applications, like drifting, this method may not work because the controller may not be able to converge to a solution fast enough to give accurate control and dynamic parameters. This lag may be solved by making the computation time faster and the convergence time faster.

The last major limitation is that user intuition is needed to determine the underlying structure of the models. If the underlying structure is selected poorly, the performance of the method may be greatly reduced. Thus, care must be taken to ensure that the underlying structure reflects the actual physics.

## VII. CONCLUSION

In this paper, we proposed a new method for learning a terramechanics model and an optimal controller using a genetic algorithm. Unlike methods used in the past research for learning terramechanics and optimal control models, the proposed method creates an interpretable model, which can be used to inform design changes of the vehicle or controller and also derive scientific conclusions.

Simulation results show that for a practical system, the proposed method performs approximately equal to a state-of-the-art NN approach, while having the benefit of producing an interpretable model. The simulation results also suggest the computation time is low enough such that the method can be implemented on a physical system with limited computational capability.

ACKNOWLEDGMENT

The authors thank undergraduate researchers Eric Langrebe, Chaska Yamane, and research adviser Ross Knepper.

## BIOGRAPHY

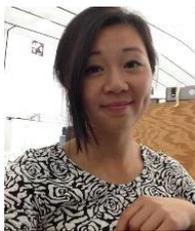

***Frances Zhu*** earned her B.S. in Mechanical and Aerospace Engineering from Cornell University, Ithaca in 2014 and is currently pursuing a Ph.D. in Aerospace Engineering at Cornell. Since 2014, she has been a Research Assistant with the Space Systems Design Studio, specializing in dynamics, systems, and controls engineering. Her research interests include flux-pinned interface applications, spacecraft system architectures, robot dynamics, estimation, and controls. Ms. Zhu is a NASA Space Technology Research Fellow.

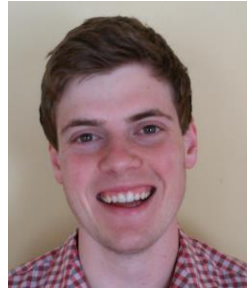

***D. Sawyer Elliott*** graduated from Rochester Institute of Technology with a Bachelor of Science in Mechanical Engineering in 2015. During his undergraduate career he worked at M.I.T. Lincoln Laboratory, where he worked on small spacecraft for weather sensing. Currently, he is a Ph.D. candidate at Cornell University under Professor Mason Peck, with a focus on dynamics and controls. While at Cornell University he worked on a small spacecraft project with the goal of autonomously docking two CubeSats. His current research explores control methods for momentum control systems, as well as the control of gyroscopically actuated robotic systems and their applications for exploration of extreme terrain.

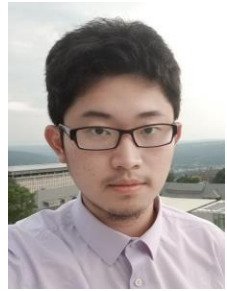

***Zhidi Yang*** received his B.S. degree in Mechanical Engineering and his B.S. degree in Computer Technology from Shanghai Jiao Tong University. He received his Master of Engineering degree in Mechanical Engineering from Cornell University. He specializes in Controls, Dynamics, and Robotics. His research focuses on locomotion, manipulation, planning, and control for legged robots, intelligent unmanned vehicles, and other high-dimensional dynamic systems. His recent work centers on applying machine learning to design of robust control algorithms.

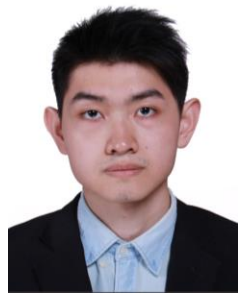

***Haoyuan Zheng*** received his B.S. degree in Electrical Engineering from Harbin Institute of Technology. He received his Master of Engineering degree in Electrical and Computer Engineering with a specialization in robotics from Cornell University. He is specialized in vision and decision making in robotics. He is interested in using machine learning methods to enhance robotic capabilities in vision and decision making.

## APPENDIX

### A. Dynamics Model

A dynamics model of the vehicle is derived to enable coarse tuning of the baseline controller, enable pre-training for the NN learning algorithm, and provide a method for evaluating each population of the genetic algorithm. The dynamics are derived for the RC car described above in Section IV but could be modified to represent a full-sized car or other vehicles.

### B. Simplifications

For the derivation of the dynamics model, seven simplifications are made:
1. The vehicle and the surface are assumed to be rigid bodies.
2. The commands for rear wheel speed and steering angle are assumed to be implemented instantaneously, thus are quasi-static parameters.
3. The wheels are assumed to have zero inertia.
4. The car is assumed to only have two wheels: one in the front of the car and one in the back. By assuming only one wheel in the front, there is no need to encode the kinematics of the Ackerman steering into the model, simplifying the derivation. Simplifying the model to have two wheels necessitates the unit vector $\hat{b}_2$ shown in Figure 9 to be constrained to be parallel with the surface, constraining the car to be upright.
5. The only contact points between the car and surface considered are the two points where the wheels contact the surface.
6. Both wheels are constrained to stay in contact with the surface at all times.
7. The friction between each wheel and the surface is assumed to be a combination of coulomb friction and viscous friction with a single $\mu$ value as shown in Eq. (21), where $V$ is the difference in velocity between the wheel and the surface that the wheel is in contact with and $F_N$ is the magnitude of the normal force due to contact between the wheel and the surface.

$$F_f = -\mu\big(V + F_N \text{sign}(V)\big) \quad (21)$$

Figures 9 and 10 show the simplified model of the RC car.

The seven simplifications above reduce the computational intensiveness of evaluating the equations of motion, enabling the GA to be implemented onto a system with limited computational capability. However, the simplified model does not accurately model the dynamics of the car on all terrains. For applications with more complex vehicle dynamics or terrain, the designer must decide on a proper model fidelity such that the dynamics accurately represent their system while remaining computationally tractable.

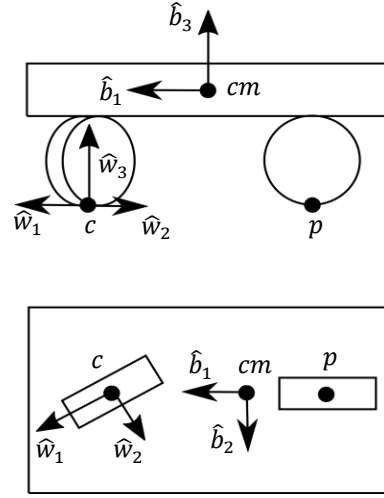

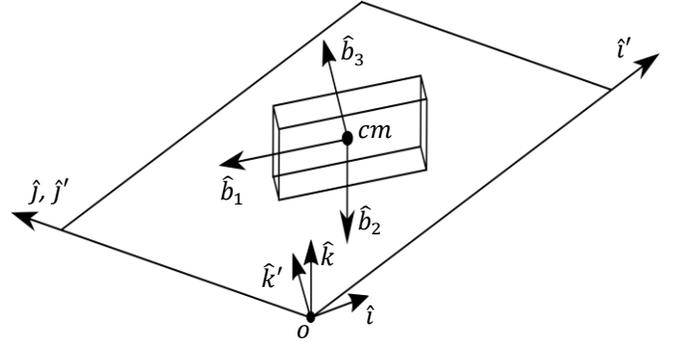

**Figure 9.** The point *cm* represents the center of mass of the car, including the wheels. The points *p* and *c* represent the points on the rear and front wheels that are in contact with the surface. The unit vectors $\hat{w}_i$ represent the coordinate system associated with the front wheel-fixed frame of reference, W. The unit vectors $\hat{b}_i$ represent the coordinate system associated with the body-fixed frame of reference, B.

**Figure 10.** The unit vectors $\hat{i}$, $\hat{j}$, and $\hat{k}$ represent the coordinate system for an inertially-fixed frame of reference, N, which is not aligned with the plane. The unit vectors $\hat{i}'$, $\hat{j}'$, and $\hat{k}'$ represent the coordinate system associated with an inertially-fixed frame of reference, N′, which is aligned with the plane. For the derivation shown, $\hat{j}$ and $\hat{j}'$ are equal. For arbitrary orientations of the slope, $\hat{j}$ and $\hat{j}'$ may not be equal.

### C. Derivation

#### 1) Nomenclature

For the derivation, the following nomenclature is used. Not all symbols in each equation are described below, but an example of each symbol is provided. The vector $r^{p/o}$ is a vector spanning from point $o$ to point $p$. The vector $\frac{^N\text{d}}{\text{d}t}r^{p/o}$ is the derivative of $r^{p/o}$ with respect to the inertially-fixed frame, N. The vector $\frac{^N\text{d}^2}{\text{d}t^2}r^{p/o}$ is the derivative of $\frac{^N\text{d}}{\text{d}t}r^{p/o}$ with respect to the inertially-fixed frame, N. The vector $\boldsymbol{\omega}^{W/B}$ is the rotation rate of frame W with respect to B. The dyadic $\underline{\boldsymbol{I}}$ is the vehicle's inertia dyadic about its center of mass. The scalar $m$ is the mass of the vehicle and $\boldsymbol{g}$ is the gravity vector.

*1) Equations of Motion*

The equations of motion are derived by first determining the equations for the rate of change of the system's angular and linear momenta with respect to the inertially-fixed frame, N. The derivative of the angular momentum of the vehicle with respect to the inertially-fixed frame, N, about point $o$ is shown in Eq. (22).

$$\frac{{}^N\mathrm{d}}{\mathrm{d}t}\boldsymbol{H} = \boldsymbol{r}^{cm/o} \times m \frac{{}^N\mathrm{d}^2}{\mathrm{d}t^2}\boldsymbol{r}^{cm/o} + \underline{\boldsymbol{I}} \cdot \frac{{}^N\mathrm{d}}{\mathrm{d}t}\boldsymbol{\omega}^{B/N} + \boldsymbol{\omega}^{B/N} \times (\underline{\boldsymbol{I}} \cdot \boldsymbol{\omega}^{B/N}) \tag{22}$$

The derivative of linear momentum of the car with respect to the inertially-fixed frame is shown in Eq. (23).

$$\frac{{}^N\mathrm{d}}{\mathrm{d}t}\boldsymbol{L} = m \frac{{}^N\mathrm{d}^2}{\mathrm{d}t^2}\boldsymbol{r}^{cm/o} \tag{23}$$

Next, equations are derived for the forces and moments acting on the vehicle. There is a total of three forces acting on the body: one contact force at each wheel, and the force due to gravity acting at the system's center of mass. The contact forces at point $p$ and $c$ are each separated into three orthogonal forces: 1) forces normal to the surface acting in the $\hat{b}_3$ direction, denoted as $\boldsymbol{F}_{p_{b_3}}$ and $\boldsymbol{F}_{c_{b_3}}$; 2) friction forces due to the commanded rotation rate of the wheels in the $\hat{b}_1$ and $\hat{w}_1$ directions, denoted as $\boldsymbol{F}_{p_{b_1}}$ and $\boldsymbol{F}_{c_{w_1}}$; 3) frictions forces due to the sideways slipping of the wheels in the $\hat{b}_2$ and $\hat{w}_2$ directions, denoted as $\boldsymbol{F}_{p_{b_2}}$ and $\boldsymbol{F}_{c_{w_2}}$. The equations describing the friction forces $\boldsymbol{F}_{p_{b_1}}$, $\boldsymbol{F}_{c_{w_1}}$, $\boldsymbol{F}_{p_{b_2}}$, $\boldsymbol{F}_{c_{w_2}}$ are shown in the Eqs. (24-31), where $r_w$ is the radius of the front and rear wheels, and $\omega_w$ is the rate of rotation of the front and rear wheels about the $\hat{w}_2$ and $\hat{b}_2$ axes, respectively.

$$\boldsymbol{F}_{p_{b_1}} = \mu_w \left(\Delta V_{p_{b_1}} + \left|\boldsymbol{F}_{p_{b_3}}\right| \text{sign}\left(\Delta V_{p_{b_1}}\right)\right) \hat{b}_1 \tag{24}$$

$$\Delta V_{p_{b_1}} = r_w \omega_w - \frac{{}^N\mathrm{d}}{\mathrm{d}t}\boldsymbol{r}^{p/o} \cdot \hat{b}_1 \tag{25}$$

$$\boldsymbol{F}_{c_{w_1}} = \mu_w \left(\Delta V_{c_{w_1}} + \left|\boldsymbol{F}_{c_{b_3}}\right| \text{sign}\left(\Delta V_{c_{w_1}}\right)\right) \hat{w}_1 \tag{26}$$

$$\Delta V_{c_{w_1}} = r_w \omega_w - \frac{{}^N\mathrm{d}}{\mathrm{d}t}\boldsymbol{r}^{c/o} \cdot \hat{w}_1 \tag{27}$$

$$\boldsymbol{F}_{p_{b_2}} = -\mu_s \left(\Delta V_{p_{b_2}} + \left|\boldsymbol{F}_{p_{b_3}}\right| \text{sign}\left(\Delta V_{p_{b_2}}\right)\right) \hat{b}_2 \tag{28}$$

$$\Delta V_{p_{b_2}} = \frac{{}^N\mathrm{d}}{\mathrm{d}t}\boldsymbol{r}^{p/o} \cdot \hat{b}_2 \tag{29}$$

$$\boldsymbol{F}_{c_{w_2}} = -\mu_s \left(\Delta V_{c_{w_2}} + \left|\boldsymbol{F}_{c_{b_3}}\right| \text{sign}\left(\Delta V_{c_{w_2}}\right)\right) \hat{w}_2 \tag{30}$$

$$\Delta V_{c_{w_2}} = \frac{{}^N\mathrm{d}}{\mathrm{d}t}\boldsymbol{r}^{c/o} \cdot \hat{w}_2 \tag{31}$$

The sum of all forces acting on the vehicle is shown in Eq. (32).

$$\boldsymbol{F}_T = \boldsymbol{F}_{p_{b_1}} + \boldsymbol{F}_{p_{b_2}} + \boldsymbol{F}_{p_{b_3}} + \boldsymbol{F}_{c_{w_1}} + \boldsymbol{F}_{c_{w_2}} + \boldsymbol{F}_{c_{b_3}} + m\boldsymbol{g} \tag{32}$$

The sum of all moments acting on the vehicle about point $o$ is shown in Eq. (33).

$$\boldsymbol{M}_T = \boldsymbol{r}^{p/o} \times \boldsymbol{F}_{p_{b_1}} + \boldsymbol{r}^{p/o} \times \boldsymbol{F}_{p_{b_2}} + \boldsymbol{r}^{p/o} \times \boldsymbol{F}_{p_{b_3}} + \boldsymbol{r}^{c/o} \times \boldsymbol{F}_{c_{w_1}} + \boldsymbol{r}^{c/o} \times \boldsymbol{F}_{c_{w_2}} + \boldsymbol{r}^{c/o} \times \boldsymbol{F}_{c_{b_3}} + \boldsymbol{r}^{cm/o} \times m\boldsymbol{g} \tag{33}$$

The conservation of angular and linear momenta yield Eqs. (34 - 35).

$$\frac{{}^N\mathrm{d}}{\mathrm{d}t}\boldsymbol{L} = \boldsymbol{F}_T \tag{34}$$

$$\frac{{}^N\mathrm{d}}{\mathrm{d}t}\boldsymbol{H} = \boldsymbol{M}_T \tag{35}$$

The unconstrained vehicle has six degrees of freedom. However, the vehicle is constrained to not fall over, restricting the vehicle to five degrees of freedom: the rotation of the vehicle about $\hat{b}_2$ and $\hat{b}_3$, and the translation of the vehicle in the $\hat{b}_1$, $\hat{b}_2$, and $\hat{b}_3$ directions. The vehicle is further constrained to three degrees of freedom, the vehicle's rotation about $\hat{b}_3$, and the vehicle's translation in the $\hat{b}_1$ and $\hat{b}_2$ directions, by constraining points $p$ and $c$ to remain in contact the surface.

The three constraints in conjunction with Eqs. (34,35) enable the equations of motion of the remaining three degrees of freedom to be derived. For brevity, the equations are not shown.